\documentclass[a4paper, 10pt, conference]{ieeeconf}      

\IEEEoverridecommandlockouts                              

\usepackage[pdftex]{graphicx}
\usepackage{multirow}
\usepackage{multicol}
\usepackage{comment}
\usepackage{breqn}
\usepackage{placeins}
\usepackage{breqn}
\usepackage{comment}

\usepackage{here}

\usepackage{mathtools,amssymb,lipsum}

\usepackage{cuted}
\title{\LARGE \bf Real-time Motion Generation and Data Augmentation for\\ Grasping Moving Objects with Dynamic Speed and Position Changes}

\author{Kenjiro Yamamoto$^{1}$ Hiroshi Ito$^{1}$, Hideyuki Ichiwara$^{1}$, Hiroki Mori$^{2}$ and Tetsuya Ogata$^{2,3}$
\thanks{$^{1}$ Kenjiro Yamamoto and Hiroshi Ito are with the Robotics Research Department, Connective Automation Innovation Center, R\&D Group, Hitachi, Ltd. {\tt\small kenjiro.yamamoto.bq@hitachi.com}}%
\thanks{$^{2}$ Hiroki Mori are with the Institute for AI and Robotics, Future Robotics Organization.}
\thanks{$^{3}$ Tetsuya Ogata is with the Faculty of Science and Engineering, Waseda University. {\tt\small ogata@waseda.jp}}}
\begin{document}
\bstctlcite{IEEEexample:BSTcontrol}

\maketitle
\thispagestyle{empty}
\pagestyle{empty}

\begin{abstract}
While deep learning enables real robots to perform complex tasks had been difficult to implement in the past, the challenge is the enormous amount of trial-and-error and motion teaching in a real environment. The manipulation of moving objects, due to their dynamic properties, requires learning a wide range of factors such as the object's position, movement speed, and grasping timing. We propose a data augmentation method for enabling a robot to grasp moving objects with different speeds and grasping timings at low cost. Specifically, the robot is taught to grasp an object moving at low speed using teleoperation, and multiple data with different speeds and grasping timings are generated by down-sampling and padding the robot sensor data in the time-series direction. By learning multiple sensor data in a time series, the robot can generate motions while adjusting the grasping timing for unlearned movement speeds and sudden speed changes. We have shown using a real robot that this data augmentation method facilitates learning the relationship between object position and velocity and enables the robot to perform robust grasping motions for unlearned positions and objects with dynamically changing positions and velocities.
\end{abstract}

\section{INTRODUCTION}
One of the challenging tasks for robots is moving-object manipulation. For a robot to perform tasks that people perform on a daily basis, such as picking up rolling objects, picking up packages on a conveyor belt, and exchanging objects with people while moving, the robot needs to recognize moving objects and predict their trajectories in real time.

In a limited environment such as a factory, the task of grasping an object on a conveyor belt moving at a constant speed can be easily performed by preparing the environment for the robot. Object recognition using deep learning has made it possible to recognize complex objects such as multiple products, indefinite posture, and mixed products \cite{redmon2015real,mahler2017dex}. However, it is necessary to manually prepare correct labels in advance for highly accurate object recognition. To achieve robust and dynamic trajectory planning for object positions and shapes, conventional robot-control technology requires a high level of expertise and huge development costs.

End-to-End robot-motion generation using deep learning enables the simultaneous acquisition of diverse object recognition and robust trajectory generation. The most attractive feature of this approach is that it can significantly reduce the development process of the object model and robot trajectory control, which previously had to be given by the designer. Deep reinforcement learning enables the acquisition of optimal behaviors that might not be conceived by humans by learning actions that maximize rewards \cite{levine2016end,yahya2017collective,gu2017deep,levine2018learning}. In particular, the motion generation using Transformer showed that more than 700 tasks \cite{brohan2022rt} and flexible object manipulations \cite{kim2022robot} can be performed if a large amount of training data is prepared in advance. However, this is difficult to apply to real-time moving-object-manipulation tasks because of the time required for inference during motion generation. The learning-from-demonstration method, which generates motions on the basis of human demonstration, requires high-quality training data to acquire generalized motions. Therefore, it is difficult for even skilled operators to teach grasping motions for objects with high-speed movement.

In this paper, we proposes a data augmentation method that realize manipulation of moving objects at low cost.
First, a robot is taught to grasp a moving object with low speed at multiple positions using teleoperation, and data augmentation is applied to the collected robot sensor data. Specifically, multiple data with different speeds and grasping timings are generated by applying down-sampling and padding processing in the time-series direction. This makes it possible to prepare training data for objects moving at high speed without advanced motion teaching techniques (which is difficult even for skilled operators). 
Next, using the augmented sensor data, a motion generation model is time-series trained to minimize the prediction error.
By learning from a variety of sensor data, the robot can generate motions in real time even when the object's position, movement speed, and timing change dynamically.
In addition, we compared the task-success rate and internal representation of (1) training only human-demonstration data (conventional method) and (2) training augmented data (proposed method).
Experimental results indicate that the proposed method can easily learn the relationship between object position and velocity, and enable a real robot to perform grasping actions robustly against untaught positions and dynamic position changes.

\begin{figure*}[th]
\begin{center}
\includegraphics[width=\textwidth]{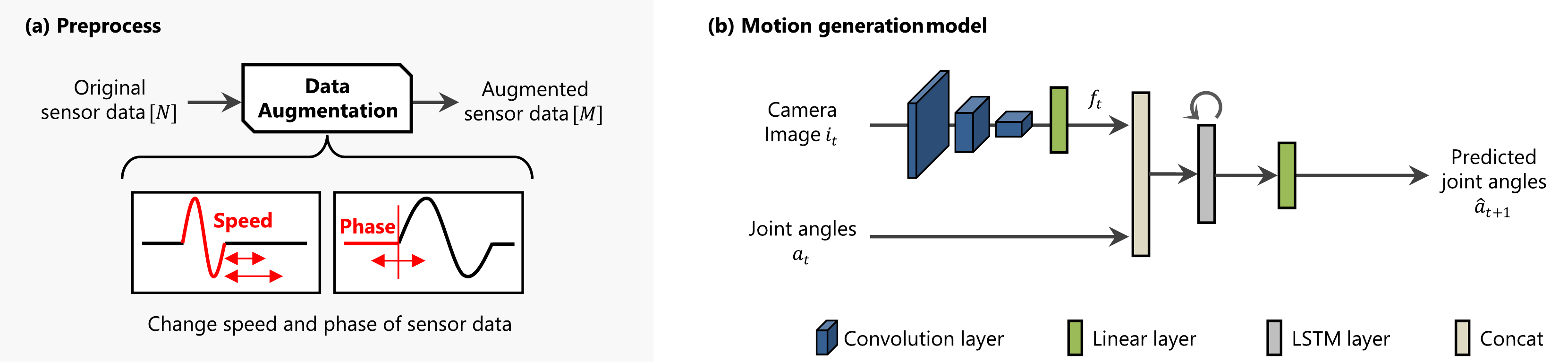}
\caption{Overview of data augmentation method and motion generation model}
\label{fig:system_overview}
\end{center}
\end{figure*}

\section{RELATED WORK}
For the study of manipulating moving objects, there are method that combine high-speed vision, rapid actuation, and visual feedback. A method was proposed to process images captured at 1000 fps through parallel processing using an field-programmable gate array, enabling the detection, recognition, tracking, as well as computation of moments and motion directions of the target object \cite{yamazaki20174}.
It has thus become possible to recognize a rapidly moving ball within 1 ms and adjust the motion based on its color. By modifying the manipulator's trajectory in accordance with the ball's movement every 1 ms, a batting task can be performed \cite{senoo2004high}\cite{senoo2006ball}. However, because of the required high-speed recognition and trajectory-planning processes, object recognition is based on the color center of gravity using dedicated hardware, which limits the number of recognizable objects.

Machine-learning-based methods have been proposed for object recognition and motion generation with generalization performance. There is method of catching moving objects in flight (balls, bottles, hammers, tennis rackets, etc.) by programming-by-demonstration \cite{kim2014catching}. This method first uses machine learning in advance to learn the trajectory prediction and optimal grasping posture of the object. Next, a human operator teaches the robot the graspable position and orientation using direct teaching. The robot then learns the optimal grasping posture and arm movement taught by the human operator to grasp the tossed object. There are also methods that use imitation learning and reinforcement learning to interact with moving objects \cite{maeda2018reinforcement}. Optimizing policies to quickly adapt to large/unexpected changes in the object makes it possible to respond to different periodic movements. However, the cost of collecting training data is an issue (limited task) with all these methods, and the realization of moving object grasping.

The authors propose "deep predictive learning" which is a cost-effective motion generation method for performing complex tasks in real time\cite{Ito2022sr}. In this method, a robot is taught a desired motion several times in the real world using teleoperation, and the model is trained using the robot sensor data at that time. By training RNN (Recurrent Neural Network) to minimize the prediction error of sensori-motor information at the current and next time step, the motion elements are modeled as attractors inside RNN. During inference, the RNN's attractor retraction action is used to generate motions in real time to minimize the prediction error, which makes it possible to realize various tasks such as handling flexible objects and door-opening motions using a real robot\cite{Koma2016ral, Ichiwara2022icra, saito2021select}.

In this paper, we describe a data augmentation method for cost-effective manipulation of moving objects using a motion generation model that performs time series prediction. We also show that the proposed method can perform grasping actions robust to the position, speed, and timing of the object by simply teaching multiple grasping actions of a slow-moving object. Specifically, it is possible to manipulate objects moving at high speeds, which is difficult to teach manually. Note that this method can be applied to a wide variety of learning algorithms and is not limited to deep predictive learning.

\section{PROPOSED METHOD}

Fig. \ref{fig:system_overview} shows an overview of the proposed method, which consists of (a) a preprocessing function that expands data from previously collected demonstration data and (b) a motion generation model that generates motions in real time using the robot's visuomotor information. Details of each component are described below.

\subsection{Data Augmentation of Motion Speed and Phase}
\begin{figure}[th]
\begin{center}
\includegraphics[width=86mm]{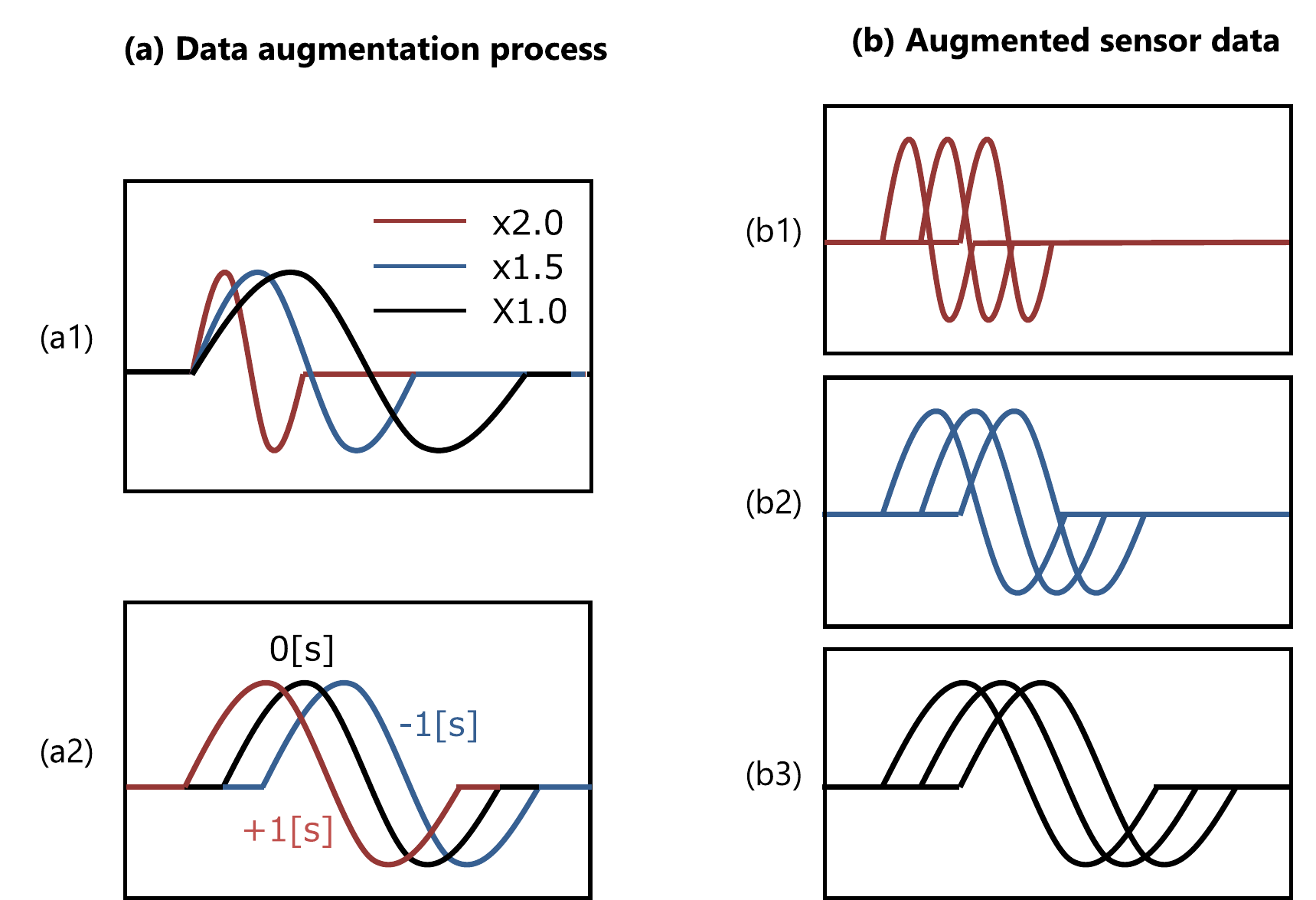}
\caption{Examples of data augmentation of speed and phase}
\label{fig:2_augmentation}
\end{center}
\end{figure}

Fig. \ref{fig:2_augmentation} shows an overview of the data augmentation processes in which (a) velocity and (b) phase (phase of object position and robot posture: timing) are changed with respect to previously collected sensor data (original data). To explain this data augmentation process clearly, a certain robot joint angle is considered as a Sin wave. Note that other joint angles and image data are processed in the same manner.

Fig. \ref{fig:2_augmentation} (a1) shows the data augmentation of velocity, in which the original data are down-sampled at a specific sampling interval to generate time-series data with different velocities. The black line in the figure is the original data, and the blue and red lines are the data with 1.5 times and 2 times the velocity ratio, respectively. For example, taking out 2 out of 3 steps of the original data yields the image and joint angle data when the object moves at 1.5 times the speed, and taking out 1 out of 2 steps of the original data yields the image and joint angle data when the object moves at twice the speed. Although augmenting the data by up-sampling the robot joint angles using linear interpolation or other methods is possible, the method of up-sampling the images is an issue. Therefore, we teach a grasping motion of an object moving at low speed and down-sample the data to generate multiple time-series data with a higher speed than the original data.

Fig. \ref{fig:2_augmentation} (b1) shows the data augmentation for phase, in which the original data are shifted in the positive/negative direction on the time axis (delay amount: -10, -5, +5, +10 steps, etc.) to generate data with different grasping timing. For example, by padding the data at time 0, phase-delayed data are generated, and by trimming a certain width of data from time 0, phase-advanced data are generated. These different timing data are effective for adjusting the predicted value of object position deviation (phase) and grasping timing in the case of sudden speed change.

By combining the above two processes, it is possible to generate a variety of data with different speeds and phases. 
Fig. \ref{fig:2_augmentation} (b1) is twice the speed of the original data, (b2) is 1.5 times the speed of the original data, and (b3) is the original data. The figure also shows an example of generating nine types of training data from the original data by changing three types of phases for each type of data.

\subsection{Motion Generation Model}
Fig. \ref{fig:system_overview} (b) shows the motion generation model used in this paper. This model learns end-to-end time-series relationships of visuomotor information when the robot interacts with the environment. The training data are not labeled with correct answers, and the model is trained to predict the robot state of the next step as an input to the current robot state. This time-series learning eliminates the need for detailed design of a physical model of the environment, which is required in conventional robotics. The model used in this paper consists of a Convolutional Neural Network (CNN) that extracts image features and a Long Short-Term Memory (LSTM) that performs time-series predictive learning. In this paper, LSTM is used for time-series learning, but other time-series learning models can be used, such as GRU\cite{cho2014learning}, MTRNN\cite{yamashita2008emergence} and Transformer\cite{vaswani2017attention}. Since the high-dimensional raw image data $i_t$ acquired by the robot contain a large amount of unnecessary information for the robot task, such as background images, the CNN is used to extract the image features $f_t$ that are important for the task. The extracted image features $f_t$ and robot joint angle $a_t$ are combined in the concat layer and input to LSTM to output (predict) the robot joint angle $\hat{a}_{t+1}$ at the next time step. During task execution, visual and motion information of the robot is sequentially input to the model, and the output (predicted) joint angles are used as motion command values to control the motors. This enables the robot to generate movements in real time in accordance with to the position and movement speed of objects.
The MSE (Mean Squared Error) is used for the loss function and Adam (Adaptive Moment Estimation) is used for the optimization algorithm\cite{kingma2014adam}.

\section{EXPERIMENTS}

\subsection{Hardware}
\begin{figure}[H]
\begin{center}
\includegraphics[width=86mm]{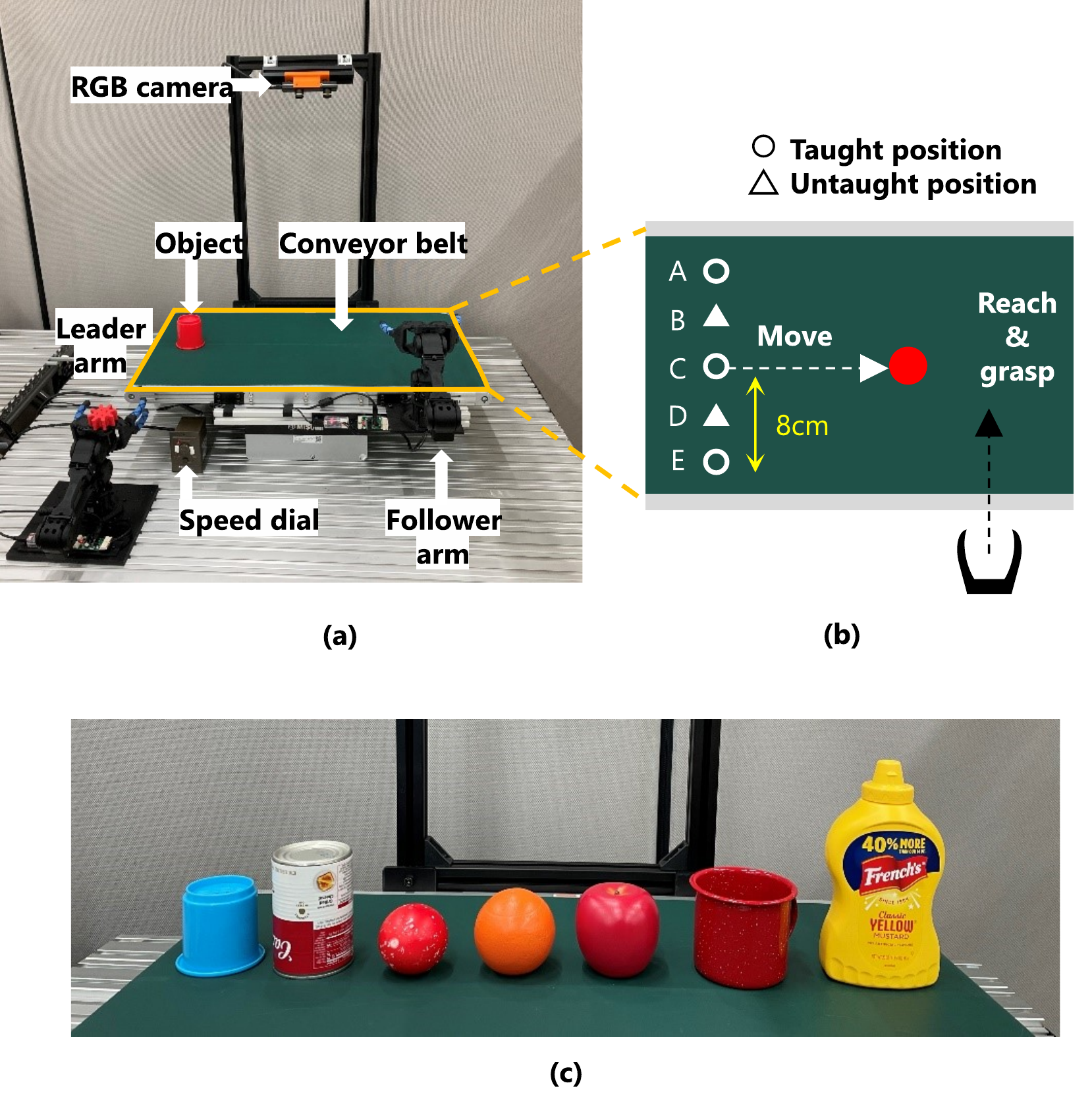}
\caption{Experimental task}
\label{fig:3_robot_system}
\end{center}
\end{figure}

Fig. \ref{fig:3_robot_system} (a) shows the experimental setup consisting of a conveyor belt, robot arms (leader and follower), camera, and target object (red cylinder). Note that the conveyor belt is completely independent of the robot arms and control PC, and no speed information is exchanged. The robot arm consists of a 4 degree of freedom (DoF) and a 1-DOF gripper, and a Dynamixel XM430 servo motor is used for each joint. 
The leader arm is operated by a human operator and its motion is transmitted to the follower arm to teach the motion. A stereo camera (ZED mini) is mounted on top of the conveyor belt to measure the states of the moving object and robot arms.

\FloatBarrier

\subsection{Task and Dataset}\label{seq:data}
We set up an experimental setting with the task of grasping an object moving from left to right on a conveyor belt. The robot is taught to (1) reach out its arm when the object approaches, (2) grasp the object, and (3) return to the initial position. To further verify generalization performance, we evaluated whether the robot can generate grasping motions even with untaught movement speeds, positions, and untaught objects.

Fig. \ref{fig:3_robot_system} (b) shows an overview of teaching motion. The robot is taught to grasp the moving object at three different initial positions (marked with circles). The speed of the conveyor belt is set to 50 rpm, which is the lowest setting, and the robot is taught to grasp the moving object once at each position using teleoperation. Camera images (64 × 128-pixels) and joint angles (5-DoF) are collected at a sampling rate of 10 Hz for 14 seconds (140 steps). Using the proposed method, a total of 27 time-series data sets are generated at the three grasping positions, combining three different velocity ratios (1.0, 1.5, and 2x) and three different delay amounts (0, +10, and +20[step]). Note that the data with a speed ratio of 1.0 and delay of 0 steps are the same as the original data. For the evaluation data, 4 types of time-series data are generated by combining the untaught speed ratios (1.25 and 1.75 times) and delay amounts (+5 and +15 steps). To verify the effectiveness of the proposed method, we manually collected 27 types of data under the same conditions as the augmented data.

Fig. \ref{fig:3_robot_system} (c) shows the target objects, which were selected from the daily items provided in the YCB dataset\cite{calli2015benchmarking} that the robot can grasp.

\section{RESULTS AND DISCUSSION}

\subsection{Moving-object Grasping Task}
\begin{figure}[htb]
\begin{center}
\includegraphics[width=86mm]{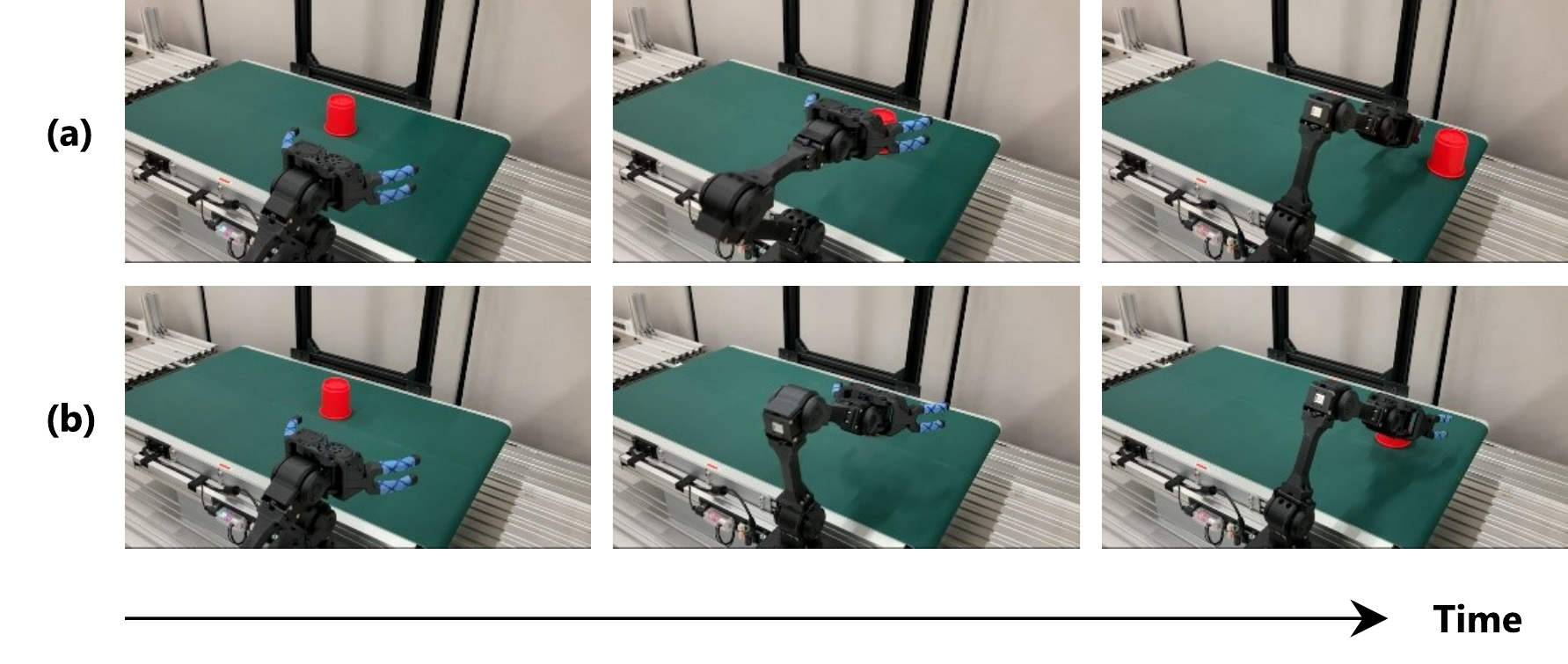}
\caption{Task-failure scene with conventional method}
\label{fig:4_task_failure}
\end{center}
\end{figure}

Table \ref{tab:results1} lists the task-success rates when the conveyor belt moved at constant and variable speeds. Success in this experiment was defined as being able to grasp the moving object properly, and 10 trials were performed for each of the experimental conditions. In the constant-speed experiment, a total of five patterns were tested: learned speeds of 50, 75, and 100 rpm and untaught speeds of 68 and 88 rpm. In the variable-speed experiment, a total of three patterns were tested: 50 to 100 rpm, 100 to 50 rpm, and random. In Table \ref{tab:results1}, rows (a) and (b) show the task-success rates of the conventional method when 27 types of data were collected manually for object-grasping motions with different positions and speeds, and rows (c) and (d) show the task-success rate of the proposed method when 3 types of data were collected manually for moving-object grasping (50 rpm) with different initial positions (3 types) and increased to 27 types of data with the proposed method. Table \ref{tab:results1} rows (a) and (c) show the task-success rates at the teaching position (C in Fig. \ref{fig:3_robot_system}(b)) with the conventional and proposed methods, respectively, and both methods had a task-success rate of 100\% with exceptions. Table \ref{tab:results1} rows (b) and (d) show the task-success rates at the untaught position (B in Fig. \ref{fig:3_robot_system}(b)) with the conventional and proposed methods, respectively. The proposed method had a task-success rate of 100\%, while the conventional method's was 0\%. Similar task-success rates were obtained for other object positions (A, D, and E in Fig. \ref{fig:3_robot_system}(b)), with 100\% task-success rate for taught positions for both methods and 0\% task-success rate for untaught positions with the conventional method. The conventional method reached near the object, but in many cases failed in the task because it could not generate movements with appropriate grasping timings. Fig. \ref{fig:4_task_failure} shows a failure scene with the conventional method, where the speed of the conveyor belt was changed from (a) 50 to 100 rpm and (b) 100 to 50 rpm. From (a), the object had already passed by when the robot's gripper closed. In (b), the object had not yet reached the grasping position when the gripper closed. Therefore, the conventional method failed in the task due to the fact that the grasping timing could not be adjusted properly at the untaught position.

\FloatBarrier
\begin{table*}[p]
\begin{center}
\caption{Success rate of speed change task [\%]}
    \begin{tabular}{l||ccccc||ccc} \hline
                            &   \multicolumn{5}{c||}{Fixed  Speed [rpm]} & \multicolumn{3}{c}{Variable Speed [rpm]}  \\
                                      &   50   &  68  &   75  &   88  &   100  &  50 to 100 & 100 to 50   & Random  \\ \hline \hline
    (a) Conventional w/ taught position     &   100  &  100 &  100  &  100  &   100  &  100       & 80          & 100     \\
    (b) Conventional w/ untaught position   &   0    &  0   &  0    &  0    &   0    &  0         &  0          & 0       \\
    (c) Proposed w/ taught position        &   100  &  100 &  100  &  100  &   100  &  100       & 100         & 100     \\
    (d) Proposed w/ untaught position      &   100  &  100 &  100  &  100  &   100  &  100       & 100         & 100     \\\hline
  \end{tabular}
 \label{tab:results1}
\end{center}
\end{table*}

\begin{figure*}[p]
\begin{center}
\includegraphics[width=\textwidth]{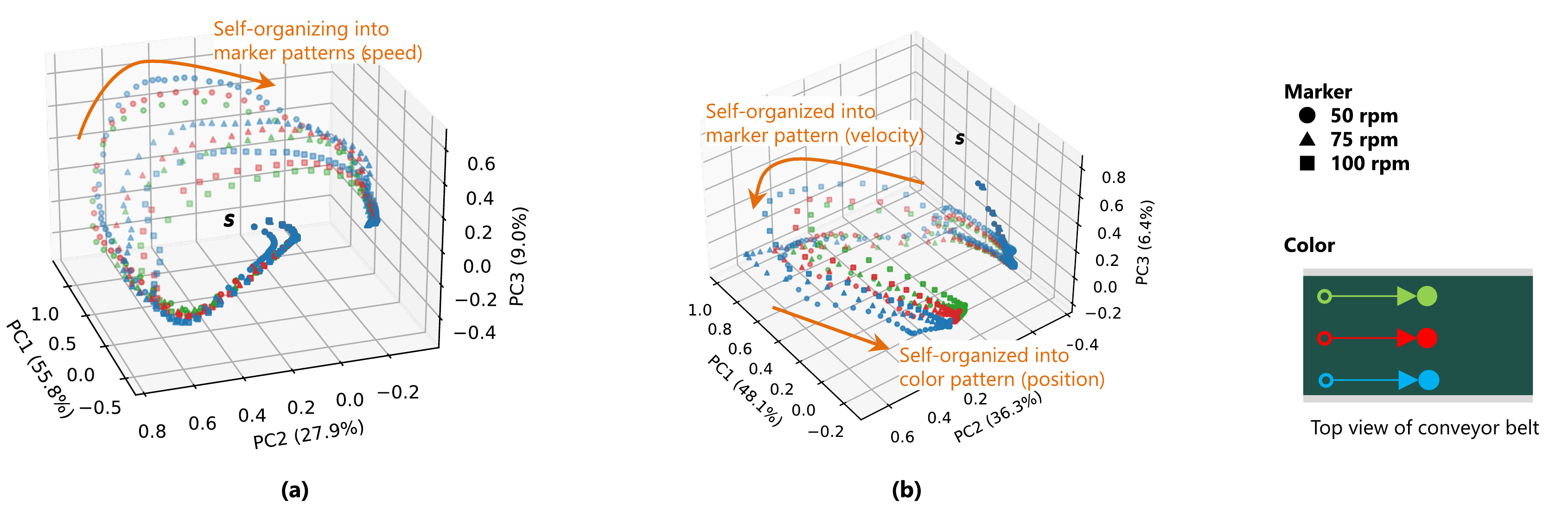}
\caption{Internal representation analysis of LSTMs}
\label{fig:5_PCA}
\end{center}
\end{figure*}

\begin{figure*}[p]
\begin{center}
\includegraphics[width=\textwidth]{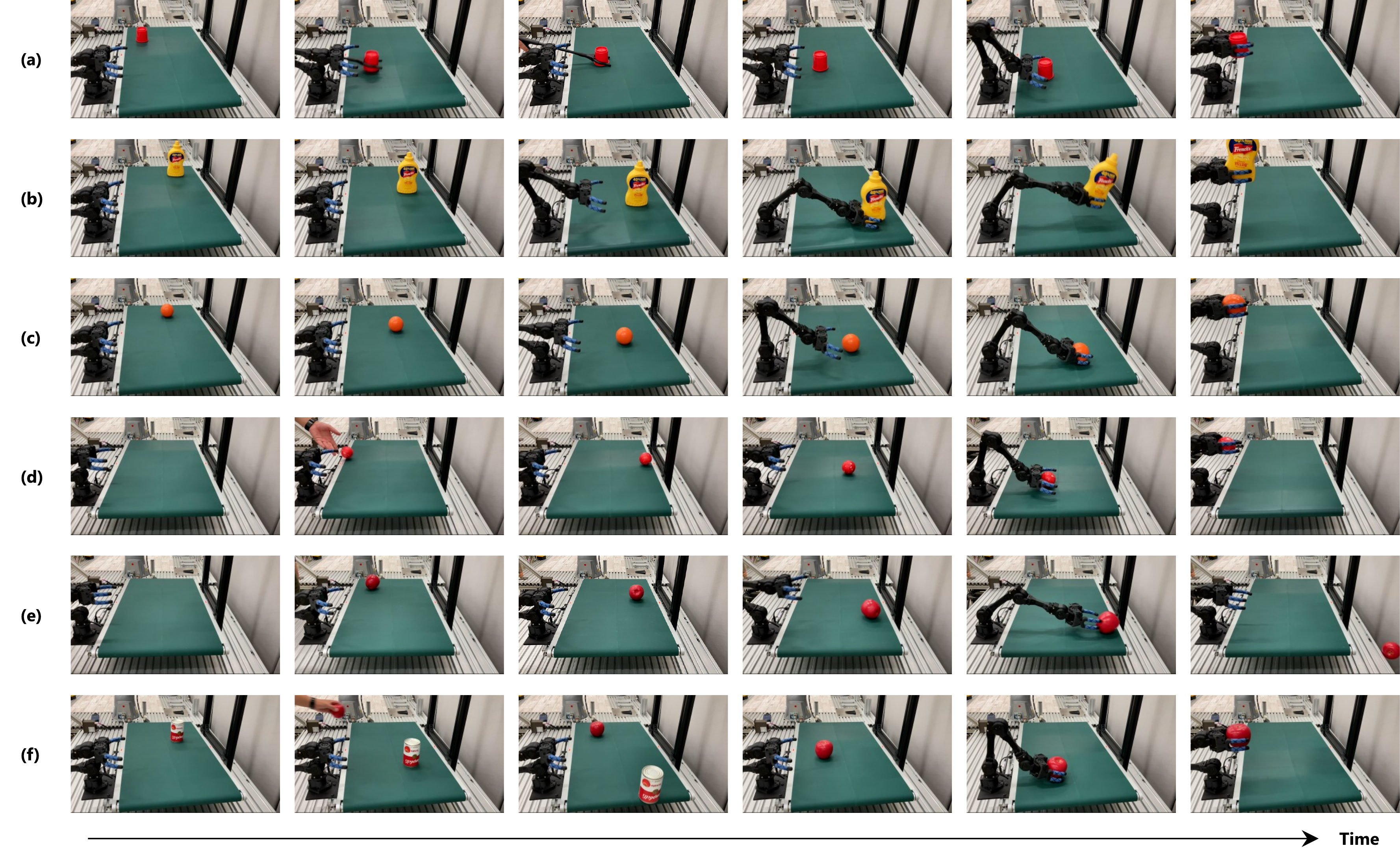}
\caption{Robustness evaluation for untaught objects and situations}
\label{fig:6_snapshot}
\end{center}
\end{figure*}

\subsection{Internal Representation Analysis}
To investigate the reason for the difference in task-success rates at untaught positions, we visualized the internal state of the LSTMs of the conventional and proposed methods  using offline analysis. The offline analysis consists of the following three steps: (1) generate a total of nine patterns with three types of velocity data (50, 75, and 100 rpm) at each of the three teaching positions using data augmentation. (2) Input nine patterns of time-series data to each motion-generation model and save the state [number of data $times$ sequence length $times$ 50 dimensions] of the LSTM at each time. (3) Compress the internal state of the LSTM into three dimensions using principal component analysis and plot them in three-dimensional space. This procedure provides a visual understanding of how the internal state of the LSTM has transitioned at each time point, i.e., how the sensorimotor information is structured (see details in \cite{Ito2022sr}). Fig. \ref{fig:5_PCA} shows the internal representation (attractor) of the LSTMs of (a) the proposed method and (b) conventional method. The rightmost part is the legend of the figure, where the markers' shapes indicate the movement speed and their colors indicate the teaching positions. Each point moves (transitions) from the starting point "S" as the time series progresses. Fig. \ref{fig:5_PCA}(a) shows that the points overlap immediately after the start of motion, the attractors gradually separate into the shape of the markers (movement speed), and the colors (position) are aligned in blue, red, and green, indicating that information on position and speed has been learned (self-organized) within the LSTM.

Fig. \ref{fig:5_PCA}(b) is the internal representation of the conventional method, which is separated by speed (marker shape) in the first half of the start and by position (marker color) in the second half, and the information of speed and position are mixed and structured inside the LSTM. The proposed method can generate training data with consistent velocity and timing by data augmentation. The conventional method's manual data collection leads to errors in trajectory and timing, preventing separate learning of velocity and position in the LSTM, resulting in a low task-success rate due to varying acceleration, deceleration, and arm posture. Thus, generalizing using only human-taught motion data is challenging.

\subsection{Disturbance Task}
Since the proposed method can enable a robot to grasp objects with a high task-success rate at both taught and untrained positions, we evaluated the robustness of the proposed method when the object position is suddenly changed as a disturbance task.

Fig. \ref{fig:6_snapshot}(a) shows the experimental results. Just before the grasped object arrived in front of the robot, the human operator changed the object position using the reaching tool. When the robot recognized that the object was gone, it returned to the initial posture and waited for the object to come again. Note that the robot was not taught the action of returning to the initial posture but was able to generate autonomous actions on the basis of the sensor information (object position) to enable continuous work. In the supplemental video, the human operator changes the object position multiple times as the object comes in front of the robot. The robot can continue generating grasping motions repeatedly until the object is grasped. The above results indicate that the robot can generate robust motions even when the position of the object to be grasped suddenly changes.

\subsection{Untaught Object-grasping Task}
Figs. \ref{fig:6_snapshot}(b) and (c) show the results of grasping-motion generation when the untaught object was moved at the untaught speed (88 rpm). The robot was able to grasp the objects properly, although the color, size, and shape of the objects were different. However, as described in Section \ref{subseq:limitation}, the blue container and silver can, which are different in hue from the training object, did not generate a reaching motion. This indicates that the motion-generation model recognizes objects of red hue as the object to be grasped.

\subsection{Dynamic Moving-object Grasping Task}
Up to this point, we have discussed moving in a "constant direction". We now discuss the task-success rate when the position (A-E in Fig. \ref{fig:3_robot_system}(b)) and speed of the object change "dynamically" by rolling the target object on the conveyor belt.

Fig. \ref{fig:6_snapshot}(d) shows a red ball rolling perpendicularly to the direction of the conveyor belt, and the robot was able to properly grasp the object as it bounced back from the wall. The supplementary video shows that the robot succeeded in the task 5 out of 6 times, generating motions by adjusting the position, velocity, and grasping timing of the object. This task is very difficult for the robot because it needs to take into account both the position and velocity of the target object at the same time. Therefore, the conventional method could not enable the robot to grasp the object "even once" in this experimental task due to its low robustness against untaught positions. The proposed method, however, could generate grasping motions with a high task-success rate for untaught speeds and positions. Therefore, we believe that the proposed method can generate motions robustly even for objects with dynamically changing positions.

\subsection{Limitations} \label{subseq:limitation}
We discuss three limitations with the proposed method. The first limitation is the grasping of irregularly moving objects. Fig. \ref{fig:6_snapshot}(e) shows the results of rolling an apple, an irregularly shaped object, as an experimental task for grasping a dynamically moving object. Unlike a ball, which moves in a straight line, an apple moves by swaying back and forth and from side to side, making it difficult to predict its exact movement path. Therefore, although the robot reached near the object, it could not grasp it properly due to subtle differences in hand position and height. This problem can be mitigated by shortening the control cycle, but this is not a fundamental solution. However, we believe that this problem can be solved by introducing uncertainty into our motion-generation model. This model was trained to minimize the error in the predicted joint angles with respect to the taught data, thus does not take into account uncertainty in the object behavior and joint angles. Therefore, we believe that it is possible to grasp irregularly shaped objects by generating movements to minimize the prediction variance using a stochastic recurrent neural network \cite{murata2013learning}, which also predicts the variance between the predicted joint angles and predicted object position.

The second limitation is the grasping of an object that has a very different hue from the taught object (red). Fig. \ref{fig:6_snapshot}(f) shows the results of moving a silver cylindrical can as an untaught-object grasping task. The robot did not recognize the object as the object to be grasped because of its very different hue from the object to be taught (red), thus it did not move at all. When a red object (apple) was placed after the silver cylindrical can, the robot grasped it properly suggesting that it recognized the object on the basis of color information. Although this feature is beneficial because the robot grasps the taught object and ignores other objects, the robot needs to explicitly learn the target object. We believe that this problem can be solved by introducing a spatial-attention mechanism \cite{Ichiwara2022icra} that explicitly extracts the location information of the object from the image information by learning black and white image information and learning data in which the hue of the image is changed.

\begin{figure}[ht]
\begin{center}
\includegraphics[width=\textwidth/2]{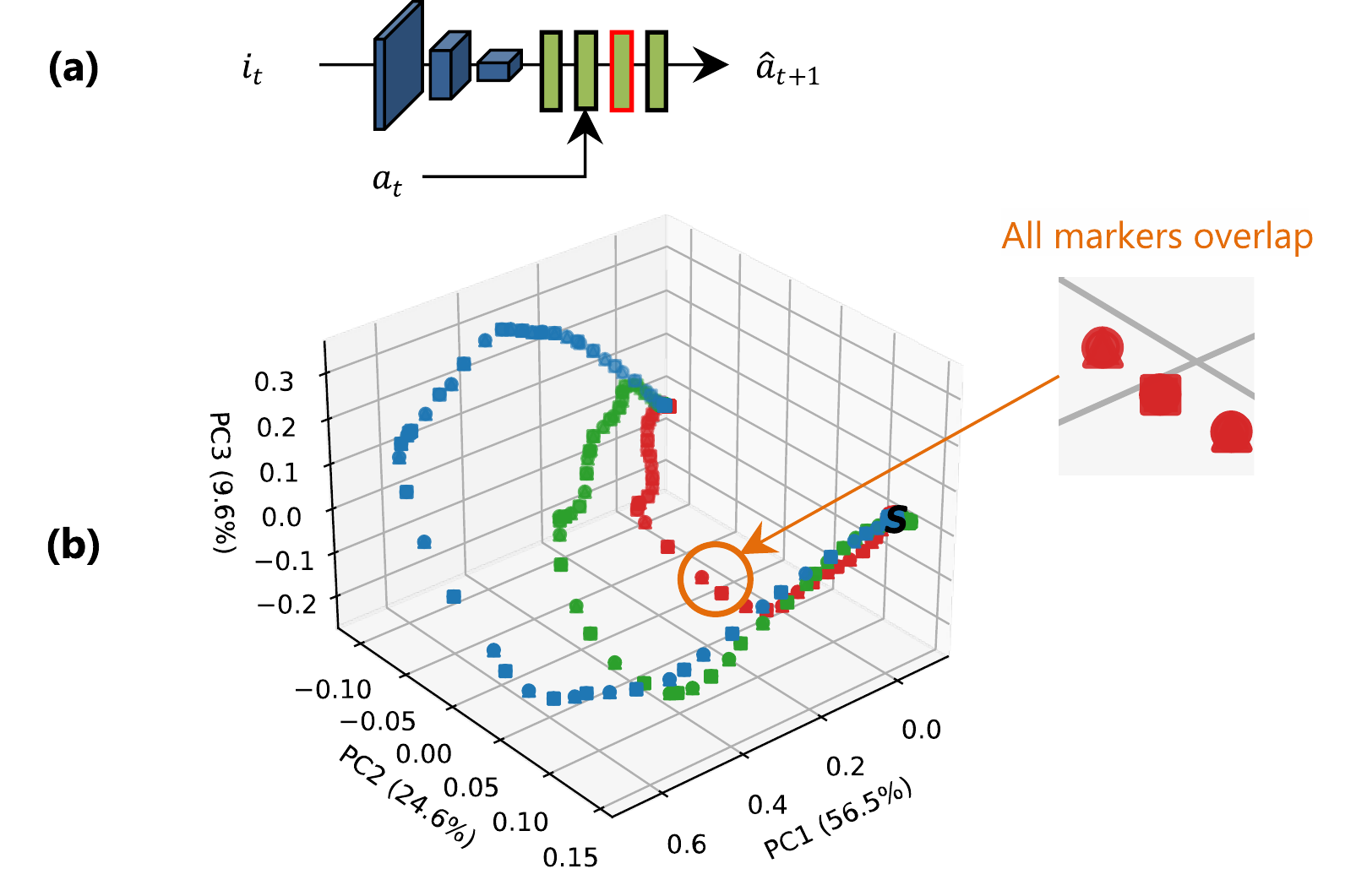}
\caption{Internal representation of motion generation model without RNN}
\label{fig:7_CNNFNN_PCA}
\end{center}
\end{figure}

The third limitation is that the proposed method is only effective for motion-generation models that learn time-series information. In this paper, we showed that an LSTM can handle untaught velocities by learning the object's position and movement speed as internal representations from the training data. We further discuss the results of applying our method to a motion-generation model without time-series learning. The training data are the same those mentioned in Section \ref{seq:data}. Fig. \ref{fig:7_CNNFNN_PCA}(a) is an overview of the model generally used in reinforcement learning \cite{levine2016end}, which predicts the action $\hat{a}^{t+1}$ at the next time step from the image $i_t$ at time $t$ and the robot's state $a_t$. Fig. \ref{fig:7_CNNFNN_PCA}(b) shows the internal representation of the test data input to the model, and the output of the neuron layer just before the action prediction (red box in Fig. \ref{fig:7_CNNFNN_PCA}(a)) is visualized by principal component analysis. The legend in the figure is the same as that in Fig. \ref{fig:5_PCA}, where the color of the marker indicates the object position and the marker indicates the movement speed. From the results of principal component analysis, it is possible to predict the motion from the image in accordance with the object position (color of the marker) since each object position (color of the marker) can be self-organized. However, the lack of self-organization by speed (all marker shapes overlap) suggests that the model may not be able to properly learn the object's movement speed. When this model was applied to a real robot, the robot was unable to grasp the moving object. The reason for this is that it is difficult for such a motion-generation model to predict the moving speed of an object from a one-time image $i_t$. To solve this problem, the model can be improved to predict the action $a_{t+1}$ at the next time step from the images $i_{t-1}$ and $i_t$ of the past few steps, as in the time-delay neural network \cite{noda2014multimodal}.

\section{CONCLUSION}
Conventional moving-object manipulation using machine learning requires learning of all factors such as the position, movement speed, and grasping timing of the target object, and collecting huge amounts of training data and conducting trial and error in the real world have been challenges. By applying the proposed data-augmentation method to the robot-sensor data obtained when a moving-object grasping motion was taught one time per teaching position using teleoperation, we confirmed that the robot can appropriately generate grasping motions for untaught speeds and sudden changes in position and speed. Although it is difficult for even an expert to teach grasping motions at high speed, the proposed method can enable a robot to grasp an object moving at high speed simply by teaching grasping motions at low speed and augmenting the data. Since the acceleration/deceleration timing and posture of the robot arm are different each time, it is difficult to generalize the conventional method using only a data set of motions taught by a human in accordance with the speed, etc. Our method has high generalization performance because it can learn consistent speed and timing (phase) data. In this study, we only used simple object grasping as an example, but we will verify the effectiveness of the proposed method for more complex tasks such as catching a ball or handing an object to another person.

\bibliographystyle{IEEEtran}
\bibliography{ref}

\end{document}